\documentclass[10pt,twocolumn,letterpaper]{article}

\usepackage{iccv}
\usepackage{times}
\usepackage{epsfig}
\usepackage{graphicx}
\usepackage{amsmath}
\usepackage{amssymb}

\usepackage[accsupp]{axessibility}  % Improves PDF readability for those with disabilities.
\usepackage{booktabs}
\usepackage{tcolorbox}
\usepackage{colortbl}
\usepackage{cite}
\usepackage{enumitem}
\usepackage{graphicx,pifont}
\usepackage[boxruled,linesnumbered]{algorithm2e}
\usepackage{wrapfig}
\usepackage{tikz}
\usepackage{pgfplots}
\usepackage{xcolor}
\usepackage{subcaption}

\usepackage[pagebackref=true,breaklinks=true,letterpaper=true,colorlinks,bookmarks=false]{hyperref}

\usepackage[capitalize]{cleveref}
\crefname{section}{Sec.}{Secs.}
\Crefname{section}{Section}{Sections}
\Crefname{table}{Table}{Tables}
\crefname{table}{Tab.}{Tabs.}

\iccvfinalcopy % *** Uncomment this line for the final submission

\def\iccvPaperID{12247} % *** Enter the ICCV Paper ID here

\ificcvfinal\fi

\begin{document}

%%%%%%%%% TITLE
\title{Federated Learning Over Images: \\ Vertical Decompositions and Pre-Trained
Backbones Are Difficult to Beat}

\author{Erdong Hu$^1$, Yuxin Tang$^1$, Anastasios Kyrillidis, Chris Jermaine \\
Rice University \\
\texttt{\{eh51, yuxin.tang, anastasios, cmj4\}@rice.edu}}

\maketitle
\ificcvfinal\thispagestyle{empty}\fi

%%%%%%%%% ABSTRACT
\begin{abstract}
   We carefully evaluate a number of algorithms for learning in a federated environment, and test their utility for a variety of image classification tasks.  We consider many issues that have not been adequately considered before: whether learning over data sets that do not have diverse sets of images affects the results; whether to use a pre-trained feature extraction ``backbone''; how to evaluate learner performance (we argue that classification accuracy is not enough), among others. Overall, across a wide variety of settings, we find that vertically decomposing a neural network seems to give the best results, and outperforms more standard reconciliation-used methods.  
\end{abstract}

\def\thefootnote{1}\footnotetext{Equal contribution.}

%%%%%%%%% BODY TEXT 

\section{Introduction}

There has been a recent influx of work aimed at developing and evaluating new algorithms for Federated Learning (FL) \cite{bonawitz2019towards, li2020federated, kairouz2021advances, chou2021efficient}, particularly for image classification \cite{fedcv, moon, zhao2018federated, hsu2020federated, chang2020synthetic, li2020fed, li2019privacy, yu2019federated, liu2020fedvision}. 
Most papers are primarily concerned with the development of innovative algorithms, and less concerned with the design of appropriate benchmarks and especially, appropriate baselines to test against. 
This paper, in contrast, is concerned with benchmarking: \textit{Which existing methods for FL work best, and under what conditions do they work, or not work?}
As such, our primary contribution is a set of carefully-designed experiments, rather than the introduction of a new FL algorithm.
Our benchmarks are designed to address the following concerns regarding the design of FL benchmarks and baselines.

% \vspace{0.2cm}
%\hspace{-0.4cm}
%\begin{minipage}[t]{0.99\linewidth} 
%\begin{tcolorbox}%[colback=gray!5,colframe=green!40!black] 
%\vspace{-0.1cm}
%\hspace{-0.4cm}

\vspace{5 pt}
\noindent
(1) \textit{Learning algorithms are typically evaluated on a few diverse data sets, rather than a large variety of more focused data sets.} %\vspace{-0.2cm}
%\end{tcolorbox}
%\end{minipage}
% \vspace{5 pt}
% \noindent 
% (1) \textit{Learning algorithms are typically evaluated on a few diverse data sets, rather than a large variety of more focused data sets.} 
The most common benchmark for evaluating federated image classification is CIFAR-100 \cite{krizhevsky2009learning}. 
This dataset includes whales, chairs, dinosaurs, and so on. 
This is a very diverse set of classes, and we are concerned that few FL tasks will involve differentiating dinosaurs from rabbits.  

For example, imagine that the members of a bird watching club taking pictures with their smartphones; club members label the pictures with the bird species, and FL is used to build a classifier. 
Or, a set of companies who are typically competitors—and hence cannot exchange data—want to work together to classify pictures of industrial drill bits based on whether they are going to fail. 
Both of these deployments involve narrow domains, and the classification problems involve fine-grained differentiation \cite{finegrained_named, learning_navigate} among members of a narrow category. 
True, not all applications of FL will be narrow, but they are likely to be far narrower than classifying whales versus chairs.  
We argue that, while evaluating on a broad data set---such as CIFAR-100---is useful, it is also necessary to evaluate FL algorithms on a variety of more narrow data sets.

%\vspace{0.2cm}
%\hspace{-0.4cm}
%\begin{minipage}[t]{0.99\linewidth} 
%\begin{tcolorbox}%[colback=gray!5,colframe=green!40!black] 
%\vspace{-0.1cm}
%\hspace{-0.4cm} \textit{2. ``Most evaluations focus on final accuracy or the number of communication rounds required for convergence.''} \vspace{-0.2cm}
%\end{tcolorbox}
%\end{minipage}

\vspace{5 pt}
\noindent
(2) \textit{Most evaluations focus on final accuracy or the number of communication rounds required for convergence.} 
\noindent FL algorithms are often evaluated by reporting final accuracies (after convergence), or by plotting test accuracy as a function of the number of epochs, or the number of communication rounds. 
However, final accuracy, at least in isolation, is not really a useful metric. After all, the simplest ``FL’’ algorithm is classical, data parallel learning \cite{localsgdconverge, use_local_sgd, parallel_sgd}. That is, run distributed gradient descent; compute gradients locally over mini-batches in a fully synchronous way, then do an all-reduce. As this is functionally equivalent to centralized learning, it is invariant to data distribution, and is likely to be the most accurate method in terms of final accuracy. However, it is inefficient in a federated environment. 

Likewise, considering accuracy as a function of communication rounds ignores the computation or communication cost of each round. Communication and computation costs can vary across methods, and are typically far more important than the number of rounds. A method that can quickly achieve high accuracy with a relatively large number of very inexpensive communication rounds---where a small fraction of the model is communicated at each round---is probably preferred to one that uses few rounds, but must transmit a huge amount of data. Similarly, for computation, given similar accuracies, an algorithm that performs one forward and one backward gradient descent pass is preferred over one that performs two additional forward passes, as extra overhead\cite{moon}, even though both operate in the same number of communication rounds. 

%\vspace{0.2cm}
%\hspace{-0.4cm}
%\begin{minipage}[t]{0.99\linewidth} 
%\begin{tcolorbox}[colback=gray!5,colframe=green!40!black] 
%\vspace{-0.1cm}
%\hspace{-0.4cm} \textit{3. ``Pre-trained feature extractors are not considered as a standard baseline.''} \vspace{-0.1cm}
%\end{tcolorbox}
%\end{minipage}

\vspace{5 pt}
\noindent
(3) \textit{Pre-trained feature extractors need to be a standard baseline.} 
\noindent Training the feature-extraction backbone that is used in most deep image processing networks---that is, the series of convolutional and pooling layers and without the final classification layer---is not easy. It is difficult to get the training process right, and training a good backbone requires a lot of computation. 

Researchers have recently suggested using pre-trained backbones, especially for few-shot training \cite{fewshot, pretrained_sythetic_deep_learning}. In a pre-trained backbone, a pre-trained feature extractor such as a convolutional neural network (CNN) \cite{cnn_intro, resnet, densenet} or vision transformer \cite{vision_transformer} is used without modification to embed an image in a high-dimensional space. Learning is then performed on the resulting vectors, and not on the original images. The goal is to leverage an existing, well-trained model on the new learning problem. Given the resource constraints in many FL scenarios, it is unclear why attempting to fully train such a backbone from scratch in a federated environment would be the first approach. Instead, we suggest simply taking a standard, pre-trained backbone (such as a ResNet \cite{resnet}, or a DenseNet \cite{densenet}, or both used together) and directly using that backbone as a feature extractor, without training.

The benefit is that each training image only needs to be pushed through the backbone one time to obtain a compact set of features, and then that set can be used during training. As the backbone need not to be communicated or repeatedly used to process the input images, both CPU/GPU cycles and communication are saved during federated training. Furthermore, final accuracies might actually be better, as this approach sidesteps the difficulty of training the backbone. %At the very least, this baseline needs to be considered in every study. 

\vspace{5 pt}
\noindent
\textit (4) \textit{Off-the-shelf performance is really what matters, but it is often neglected in research studies.} 
In a centralized environment, it is feasible to train and re-train many times, as various parameters  (learning rate, proximal weight, exact neural architecture, etc.) are tuned. In FL, this is much less feasible. Most FL scenarios imagine a resource-constrained environment (possibly involving edge devices), where one cannot ignore the cost of tuning. FL is typically happening on someone else’s hardware, and so running an algorithm many times during parameter tuning is likely to make participation in training far less palatable. 

Further, application-specific tuning may simply be impossible. A device may enter the federated training and then drop out shortly after, never to be seen again. One cannot access a missing device’s data to perform validation, and clearly, one cannot retrain over data that cannot be accessed. 

As such, we argue that ``off-the-shelf’’ performance is more important than in centralized learning. That is, it becomes important to settle on one set of universal parameters that tend to work in most deployments, such that the learning algorithm can be universally deployed without further tuning. At the very least, this is an important use case that should be covered in most experiments.

%\vspace{5 pt}
%\noindent
%\textbf{Our proposal.} In this paper, we argue for a very simple algorithm for federated image classification, called ``vertical decomposition over a pretrained backbone'' or \textsc{Verdove} for short. \textsc{Verdove} is designed with the aforementioned considerations in mind. We provide some strong evidence that this baseline is difficult to beat under realistic scenarios.  
%
%In \textsc{Verdove}, a fully-connected network is decomposed vertically into independent subnetworks \cite{ist}, and these vertical subnets are sent to sites for local training. All of this is done over feature vectors extracted using one or more pre-trained backbones.\footnote{We advocate using two, quite different backbones, for diversity.} 
%
%\textsc{Verdove} is especially attractive in the FL context.  We find that classical, full backpropogation---where the feature-extraction backbone is trained as well as the classifier---seems to have serious problems with accuracy when trained on a less diverse data set, and a pre-trained backbone works favorably.  Further, the pre-trained backbones seem to obviate the need for tuning.
%
%Since only narrow subnets are ever communicated to the various sites during federated training of \textsc{Verdove}, communication costs are very low. Further, local training happens only over a narrow subnetwork---and the feature extractor is fixed and need only be run one time per image; thus, CPU costs are also very low. 
%

\vspace{5 pt}
\noindent
\textbf{Our Contributions.}  Our contributions are as follows. \vspace{-0.15cm}
\begin{itemize}[leftmargin=*]
    \item We suggest several rules-of-thumb governing benchmark and baseline development for FL over images, and use those rules to devise an extensive FL image classification benchmark, and use that benchmark. \vspace{-0.15cm}

    \item  We show that using pre-trained features and other model reduction tools are necessary for practical FL. 

    \item We highlight some surprising behaviors of current FL algorithms, including the effect of client population size, that to our knowledge are not sufficiently explored in current literature. 

    \item Our source code is publicly available at \url{https://github.com/huerdong/FedVert-Experiments}
    
    %\item Based on the above rules, we devise a very simple FL algorithm---\textsc{Verdove}---which, by design, does well in a resource constrained environment, where CPU cycles and communication are precious and the classification task is difficult, in the sense that involves differentiating between similar objects. \vspace{-0.15cm}

    %\item Our benchmark shows that \textsc{Verdove} may be quite difficult to beat, even for state-of-the-art methods.  \vspace{-0.15cm}

    %\item Additional experiments explain why the vertical decompositions or ``subnets'' used by \textsc{Verdove} are particularly useful for FL over data that are skewed, in the sense that only a few classes reside at each site. \vspace{-0.15cm}
\end{itemize}

\section{Background}

The common setting for FL is the following: There are total $N$ devices participating in the optimization. To protect data privacy (or due to communication constraints), devices are not allowed to share local data samples to others. Each device has its own local objective function $f_i$ parameterized by its own model $w_i$. The coefficient $p_i = n_i/n$ is the number of data samples $n_i$ on each device, divided by the unified total data sample $n = \sum_{i=1}^{N}n_i$. There will be a central server that collects all the parameters $w_i$ uploaded by each device and aggregate them into a single model $w$.

The common formulation of FL is to optimize a global optimization function $L(w)$: \vspace{-0.25cm}
\begin{equation} \label{eqn:fed_obj}
\min\limits_{w} L(w) = \sum_{i=1}^{N} p_i L(w_i;\{x_{\text{loc}}, y_{\text{loc}}\}), \\[-4pt] 
\end{equation}
where $L(w_i;\{x_{\text{loc}}, y_{\text{loc}}\})$ represents the local objective for device $i$, based on local data $\{x_{\text{loc}}, y_{\text{loc}}\}$.
The natural algorithm that arises from this objective is FedAvg \cite{fedavg} which directly applies stochastic gradient descent (SGD) to learn $w_i$'s. In the most basic case, where FedAvg performs only one local gradient step, the above formulation becomes equivalent to data parallel SGD \cite{localsgdconverge, use_local_sgd, parallel_sgd}. %This algorithm performs well in the ideal scenario where client data is identically, independently distributed (i.i.d.), client devices have unlimited compute power and memory, and client-server communication has no overhead. 

%However, i
In real-world applications of FL, local data distributions are often skewed towards specific classes: i.e., 
%We can imagine a simple application of FL where citizen scientists take photographs with their mobile phones of birds to train a bird species classifier. In this scenario, the range of bird species that a particular client might encounter is limited by time of day, location, and population of different species. Consequently, 
the natural distribution of real data can be very skewed, with certain classes only existing in a very small segment of clients' devices. 
% \begin{figure}
%     \centering
%     \includegraphics[width=0.4\textwidth]{l2_norm.pdf}
%     \caption{\label{fig:l2_norm} $\ell_2$-norm distance from FedAvg-trained models parameters $w$ to a simple, vanilla centralized trained model parameters $w^{*}$ under both Non-I.I.D and I.I.D setting.}
% \end{figure}
% \begin{figure}
%   \centering
%   \begin{subfigure}{0.49\linewidth}
%   \centering
%     \includegraphics[width=1\textwidth]{Variance.pdf}
%     \caption{variance}
%     \label{fig:variance}
%   \end{subfigure}
%   \hfill
%   \begin{subfigure}{0.49\linewidth}
%   \centering
%     \includegraphics[width=1\textwidth]{Std_Error.pdf}
%     \caption{standard\_error}
%     \label{fig:std_error}
%   \end{subfigure}
%   \caption{The variance and standard error of FedAvg-trained model parameters (input layer and hidden layer) based on two different datasets: Non-I.I.D. and I.I.D. Non-I.I.D datasets will have magnitude higher variance and standard error compared to I.I.D datasets.}
%   \label{fig:var_stderr}
% \end{figure}
In these heavily non-i.i.d. scenarios, the learned $w_{i}$ in ~\cref{eqn:fed_obj} can diverge in drastically different directions and, in the most extreme cases, diverge in opposite directions. 
When FL is performed in this scenario, the averaging can result in a model drastically different from any of the client models and has degraded performance as a consequence; a phenomenon often called client drift \cite{fed_noniid}. It is this concern that has motivated the development of a number of FL methods beyond FedAvg \cite{fedprox, moon, fednova, scaffold}, a few of which will be described in the next section.

\iffalse
\begin{table}[t]
\centering
    
    \setlength\tabcolsep{3.2pt}
    \input{tables/method_comparison}  
    \caption{\textcolor{red}{Summary of different Federated Learning algorithms and their characteristics. $m$ is the size of the centrally shared model used in (stochastic) gradient descent, $t$ is the number of local training rounds, and $n$ is the number of clients.}}
    \label{fig:methods_compare_table}
\end{table}
\fi

\section{Federated Learning Methods Tested}

\subsection{The Methods}

In our experiments, we consider the following FL methods, chosen as a representative set of the state-of-the-art. %%\textcolor{red}{We primarily consider two classes of FL algorithms - averaging and decomposition. Averaging algorithms communicate the full central model to clients and use some averaging (weighted or unweighted) to combine the model. Decomposition methods send only part of the full model to clients and combine the client models by concatenating their results at the end of the communication round. Additionally, a common practice is for FL algorithms to employ some form of a proximal term. The code for these experiments are publicly available\footnote{\url{https://github.com/huerdong/VERDOVE-Experiments}}.} 

%%\textcolor{red}{Since all methods considered are derived from the stochastic gradient algorithm, their computation and communication costs depend on the size of the base model. In the case of the single hidden layer MLP used in our experiments, these costs are directly proportional to the size of the model. For an MLP with respectively input, hidden, and output dimensions $n_i$, $n_h$, and $n_o$, the size of the model is $n_i n_h+n_h n_o$. We additionally express their costs in terms of the batch size $b$, and the number of clients $c$. We give the communication cost as the number of bytes of communicating to all users for a single communication round, while the computation cost i. Both these expressions have a constant multiplicative factor common to all algorithms which we omit for conciseness. Because averaging algorithms require communicating and computing on the full base model at each client, their costs are . However, we note that in practice averaging algorithms "do more" per communication round and the }  

%%\textcolor{red}{A summary of the methods is provided in Table \ref{fig:methods_compare_table}.}

\vspace{5 pt}
\noindent \textbf{FedAvg.}
Federated averaging \cite{fedavg} is the most simple averaging algorithm and is the basis of the others, whereby client models are trained locally by (stochastic) gradient descent. These local models are aggregated by directly averaging their weights, or equivalently averaging their computed gradients using this averaged gradient for one global round of gradient descent. 

\vspace{5 pt} \noindent \textbf{FedProx.}
FedProx \cite{fedprox} is a modification on FedAvg that introduces an additional $\ell_2$-norm term to the local loss function, that takes the difference between the previously communicated global model and the current model. The aim is to avoid client drift by biasing the training towards an established model. Again, the clients then compute local models whose weights are averaged. Given the model weights $w_0$ from the previous communication round, the FedProx modification to the FedAvg objective is:
\begin{align}
    \min_{w_i} \left\{L(w_i; \{x_{\text{loc}}, y_{\text{loc}}\}) + \tfrac{\mu}{2}\| w_i-w_0\|^2_2\right\}, ~~i\in [N] \nonumber
\end{align}
where $\mu$ is a regularization parameter.

\vspace{5 pt} \noindent \textbf{MOON.}
MOON (Model-Contrastive) \cite{moon} uses a similar principle as FedProx by comparing the current model with the previous model. However, it utilizes a cross-entropy loss, rather than $\ell_2$-norm loss, as in: 
\begin{align}
    \min_{w_i} \left\{L(w_i; \{x_{\text{loc}}, y_{\text{loc}}\}) - \mu\log \tfrac{e^{\textrm{sim}(z, z_{\textrm{glob}})/\tau}}{e^{\textrm{sim}(z, z_{\textrm{glob}})/\tau} + e^{\textrm{sim}(z, z_{\textrm{prev}})/\tau}}\right\} \nonumber
\end{align}
where $\textrm{sim}()$ is the cosine similarity function, and $z, z_{glob}, z_{prev}$ are representations of the data generated by the respective model (respectively: learned, current global, previous local). $\tau$ is a temperature hyperparameter. 

% The additional term is interpreted as a contrastive loss that treats the current global model as a positive example and previous models as negative examples. 

\vspace{5 pt} \noindent \textbf{FedAdam.}
While the above algorithms modify the local training process of FedAvg, FedAdam \cite{fedopt} modifies the aggregation step. Once the averaged weights are computed via local gradient descent and their average is computed, it is used to compute a global gradient surrogate to be used in the Adam algorithm \cite{adam}. 

\vspace{5 pt} \noindent \textbf{FedNova.}
FedNova \cite{fednova} introduces a client gradient re-weighting scheme when accumulating the models parameters/gradients and generalizes FedAvg and FedProx as specific parameterizations. However, we use the specific formulation specified in ~\cite{fednova}, where these weights are normalizations of the local gradients.  

\vspace{5 pt} \noindent \textbf{IST.} We consider the previous set of learning algorithms to be ``reconciliation-used''; in that, they attempt to train a copy of the model at each site.  In contrast, IST, or \emph{independent subnet training} \cite{ist} 
does not utilize averaging or some other method to try to reconcile versions of the model trained at the various site.  Instead, IST decomposes the model and sends non-overlapping parts of the model to different sites.  In one training round, all neurons (or activations) in a neural network are randomly partitioned to the active sites. A weight is only sent to a site if it connects two neurons that have been assigned to the site.  Thus, each site is assigned only a very small subnetwork, which is trained locally for a number of gradient descent steps. At the end of local training, the updated weights are shuffled (with or without the aid of a central server) and the process is repeated.  Since subnetworks are independent, there is no averaging or other form of reconciliation needed.  There are many variants of this approach in the literature, including FjORD \cite{fjord}, HeteroFL \cite{diaoheterofl}, LotteryFL \cite{lotteryfl}, FedSelect \cite{charles2022federated}, FedRolex \cite{alam2022fedrolex}, Federated Dropout \cite{cheng2022does}, PVT \cite{yang2022partial}, and the independent subnetwork training approach in the non-FL setting \cite{ist, dun2022resist, wolfe2021gist, wang2022loft, masked_neurons}.  It is this latter version of the approach (for MLPs) that we apply here.

\vspace{5 pt} \noindent \textbf{ISTProx.}  This is IST, but with a proximal term added to the loss function.

\subsection{Analytic Comparison}

The focus of our experiments will be on comparing each of the methods based on their ability to produce high test accuracy with a low cost.  In the paper, we will define cost in terms of floating point operations (FLOPs) required to reach a given accuracy, and in terms of floating point numbers transferred. As such, we begin by considering the number of FLOPS required and the floating point numbers that must be transferred to complete one gradient descent step for each of the methods.

In our analysis, we assume that we are using a pre-trained backbone to transform each input image into a vector, and we are learning a MLP with two weight matrices. The MLP accepts an $n_1$-dimensional input. It has a hidden layer of $n_2$ neurons, and has $n_3$ outputs. Assume a batch size of $b$ at each of $s$ sites. Since each image only goes through the pre-trained backbone once, the FLOP cost of the pre-trained backbone is negligible when amortized over many gradient descent steps. We will consider only the FLOPs associated with the weights, as the cost associated with updating weights dominates the cost associated with the activation layers.

Begin with FedAvg. In this case, to push a batch through, we have approximately $2b (n_1 n_2 + n_2 n_3)$ FLOPs for the forward pass and approximately $2b (n_1 n_2 + 2 n_2 n_3)$ for the backward pass, for a total of $4 n_1 n_2 b + 6 n_2 n_3 b$ FLOPs at each site. Coordination requires $2s (n_1 n_2 + n_2 n_3)$ floating point numbers be sent (the model needs to be distributed to all sites, and then the updated model needs to be collected from all sites). The number of FLOPs incurred during averaging is typically not significant, as we are simply average $s$ copies of the two weight matrices, having $n_1 n_2$ and $n_2 n_3$ floating point numbers each, and this may be done using a distributed algorithm.  

Adding a proximal term in the case of FedProx may have little effect, as this changes the backward pass at each site by adding FLOP computations that scales linearly with the weight matrices, but does not scale with the batch size $b$. Thus, for any reasonable batch size, it is insignificant.  If multiple gradient update steps are performed at each site, it also requires storing an older copy of the weight matrices at each site.

FedAdam is a federated version of the Adam algorithm and as such, it requires storing four additional matrices---typically at a coordinator---two of which are associated with each of the MLP's weight matrices. It must approximate the gradient and run the Adam algorithm at the coordinator during each update. However, this does not affect the memory, computation, or communication requirements at each site, and so any differences may not be of practical importance.

FedNova requires a normalization step at each site during the local gradient computation, but this only requires multiplying each gradient by a scalar, which is likely insignificant and so it is also, for practical purposes, computationally no more expensive than FedAvg.
MOON is more expensive than FedAvg in terms of FLOPs, as it needs to compute multiple representations of each training iteration.  The obvious MOON implementation will require $4b n_1 n_2$ additional FLOPs compared to FedAvg at each site. In addition, MOON needs to locally maintain two additional copies of the weight matrices.  

Next, consider IST. At each site, $n_2$ is effectively cut by a fraction $s$, because the neurons in the hidden later is partitioned $s$ ways and only the weights associated with neurons assigned to a site are actually sent or used there. Hence, the total number of FLOPs is $(4 n_1 n_2 b + 6 n_2 n_3 b)/s$, and coordination requires that only $2 (n_1 n_2 + n_2 n_3)$ floating point numbers be sent and received; each weight is assigned to exactly one site. There are no FLOPs associated with coordination, as there is no averaging done; weights are simply updated and sent around. Due to the partitioning, the memory usage at each site is on average smaller than that of FedAvg by a factor of $s$. Like FedAvg, adding a proximal term has little effect on either FLOPs or bytes transferred. 

\section{Experimental Setup}

\subsection{Pre-Trained Backbones}

All of the methods we tested (except for one baseline) utilize pre-trained backbones. That is, we take a pre-trained feature image classifiers: ResNet101 \cite{resnet} and DenseNet121 \cite{densenet}, and remove the classification layers to produce two ``backbones.''  When processing an image, we simply concatenate the feature vectors produced by the two backbones to produce a 3072 dimensional vector.  We also evaluate FedProx, but without a pre-trained backbone. This version of FedProx (called ``FedFull'') uses a ResNet18 instead of a single-hidden layer MLP.

\begin{table}[t]
\centering
    
    \setlength\tabcolsep{3.2pt}
    
%\begin{table}
%\caption{Final Accuracies  for 10\% participation (\%) of 100 devices, IID data}
\begin{tabular}{cccccccc}
\multicolumn{7}{c}{(a) Final accuracies} \\
\hline
   & Birds & Cars & Flwrs & Aircrft & Textre & CFAR \\ 
       \hline
FedAvg    
    & 73.5 & 66.7 & 95.9 & 48.8 & 74.3  & \textbf{76.7}\\
    %\hline
FedProx
    & 74.0 & 65.7 & 95.8 & 48.9 & \textbf{74.7} & 75.1\\
    %\hline
MOON    
    & 74.3 &  66.0 & 95.6 & 48.9 & 74.6 & 75.3\\
FedAdam    
    & 67.6 & 51.1 & 92.4     & 41.9 & 72.8    & 70.1   \\
FedNova    
    & 73.3 & 66.8 & 95.6 & 49.0 & \textbf{74.7}    & 74.7   \\
FedFull    
    & 5.2 & 2.8 & 35.1     & 5.4 & 40.3    & 7.8   \\
IST    
    & 74.6 & \textbf{67.8} & 96.4 & \textbf{50.6} & 74.6 & 76.2\\
    %\hline
ISTProx
    & \textbf{74.8} &  66.8 & \textbf{96.5} & 50.4 & \textbf{74.7} & 76.1\\
       \hline
\end{tabular}

    \vspace{10 pt}
    %\caption{Communication (In GB) to reach $90\%$ peak accuracy with 10\% participation of 100 devices, IID data}
\begin{tabular}{cccccccc}
\multicolumn{7}{c}{(b) Communication (GB) to threshold acc.} \\
\hline
     & Birds & Cars & Flwrs & Aircrft & Textre & CFAR \\ 
       \hline
FedAvg    
     & 9 & \textbf{6} & 4 & 13 & 3 & 10 \\
    %\hline
FedProx
    & 13 & FAIL & 6 & 21 & 5 & 15\\
     %  \hline
MOON    
    & 13 & 43 & 6    & 21       & 4 & 16  \\
FedAdam    
    & 8 & FAIL & 2     & FAIL & \textbf{1}    & 7   \\
FedNova    
    & 57 & 138 & 17     & 84 & 12    & 68   \\
FedFull     
    & FAIL & FAIL & FAIL     & FAIL & FAIL    & FAIL   \\
IST    
    & \textbf{3} & 288 & \textbf{1}      & \textbf{3}      & \textbf{1} & \textbf{3}  \\
    % \hline
ISTProx
    & \textbf{3} & FAIL & 2      & \textbf{3}    & \textbf{1} & \textbf{3}\\
       \hline
\end{tabular}

    \vspace{10 pt}

%\caption{GFLOPs to reach $90\%$ peak accuracy with 10\% participation of 100 devices, IID data}
\begin{tabular}{cccccccc}
\multicolumn{7}{c}{(c) GFLOPs to threshold acc.} \\
\hline
       & Birds & Cars & Flwrs & Aircrft & Textre & CFAR \\ 
       \hline
FedAvg    
    & 904 & \textbf{642} & 445 & 1334 & 275 & 979 \\
    %\hline
FedProx
    & 53 & FAIL & 23  & 86 & 18 & 62\\
     %  \hline
MOON    
    & 110 & FAIL & 46      & 178 & 34 & 121\\
FedAdam    
    & FAIL & FAIL & 170     & FAIL & 152    & 750   \\
FedNova    
    & 1140 & 2768  & 333     & 1680 & 232    & 1355   \\
FedFull     
    & FAIL & FAIL & FAIL     & FAIL & FAIL    & FAIL   \\
IST    
    & \textbf{11} & 1152 & \textbf{6} & \textbf{14} & \textbf{3}  & \textbf{12}       \\
    %\hline
ISTProx
    & \textbf{11} & FAIL & \textbf{6} & \textbf{14} & \textbf{3}  & 13      \\
       \hline
\end{tabular}

    \caption{i.i.d. results, 100 sites, 10\% participation.}
    \label{fig-iid}
\end{table}

\subsection{Data Sets and Hyperparameters}

Overall, we conduct experiments on a diverse and broad range of image classification data sets: (1) CUB-200-2011 (Birds) \cite{birds},  (2) Stanford Cars (Cars) \cite{cars}, (3) VGGFlowers (Flwrs) \cite{flowers}, (4) Aircraft (Aircrft) \cite{aircraft}, (5) Describable Textures (Textre) \cite{dtextures}, and (6) CIFAR 100 (CFAR) \cite{cifar}. The only processing we apply on these data sets, besides using the pre-trained feature extractors, is the standard cropping and resizing of image data. For the Aircrft data set, there are multiple levels of specificity of labels - manufacturer, family, variant - we chose variant for the finest-grained classes. 

We use the Birds data set to tune the  methods. After extensive experimentation on Birds, we chose a learning rate of $0.01$, as we found this to give stable convergence in all algorithms, except FedAdam, where we choose a learning rate of $0.03$.  
For all algorithms except for IST, ISTProx, and FedFull, tuning led us to a single-hidden-layer, multi-layer perceptron (MLP) with 1,000 hidden neurons sitting on top of the feature extractor. FedFull was trained with its ``out of the box'' configuration. For IST and ISTProx, we find scaling the hidden neurons based on the number of concurrent training sites made sense (3000 neurons for ten sites, 6000 for 20, 9000 for 30, 18000 for 60, and so on). This ensures that the client models are not too small.

A number of other hyperparameters were also optimized using the Birds data set. For the various hyperparameters in each algorithm, we applied a grid search on a set list of parameter values, as described in the next subsection.
We consider the number of local iterations of stochastic gradient descent as a hyperparameter with either 1, 5, or 25 local training batches, before a communication round.  We use a batch size of 32 images.
For FedProx, FedNova, and ISTProx, we tuned the additional proximal hyperparameter $\mu$. For MOON, we have a similar weight parameter $\mu$ (the coefficient of the contrastive term) and $\tau$, the temperature of the contrastive term. FedAdam has a central learning rate as well, which we set to $0.01$.  Exact values for all of our hyperparameters are given in the appendix.
We stress that, while we performed careful tuning on the Birds data set, all hyperparameters were re-used blindly and without modification, on all other data sets.

  \begin{figure}
  \centering
  \includegraphics[width=0.45\textwidth]{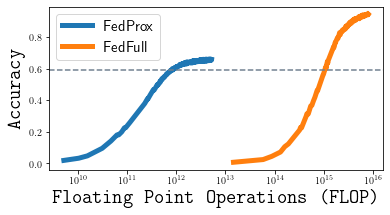}
  \caption{Accuracy as a function of time for FedProx and FedFull on the Stanford Cars data set.}
  \label{fig:fedprox_full_vs_pre}
\end{figure}

\subsection{Evaluation Metrics}

Choosing an evaluation metric is among the most difficult aspects of evaluating FL methods, but does not seem to have received much attention in the literature. Most papers focus on final accuracy: \textit{What is the accuracy that a FL method exhibits on a test set after training?} Unfortunately, final accuracy as a metric makes little sense in isolation from other considerations.  After all, the method that likely gives the very best final accuracy is simple distributed mini-batch training, where each site computes gradients over a small subset of its local data; those gradients are summed and a central site updates the model.  This is algorithmically equivalent to centralized training, but it is typically not tested, because it is generally assumed that the convergence would be too ``slow,'' for some implicit definition of ``slow.''  Thus, a more carefully-defined evaluation framework (beyond simple accuracy) is needed.

Hence, we consider three metrics: \emph{final accuracy, computational efficiency,} and \emph{communication efficiency} \cite{trading_comp_comm, rfid}.  Of the three, we assert that final accuracy should be considered the \emph{least important}; it is most useful as a way to contextualize the other two metrics (e.g., ``\textit{what final accuracy can I achieve in $G$ gigaFLOPs of computation?}'').

\begin{table}[t]
\centering
    
    \setlength\tabcolsep{3.2pt}
    
\begin{tabular}{cccccccc}
\multicolumn{7}{c}{(a) Final accuracies} \\
\hline
   & Birds & Cars & Flwrs & Aircrft & Textre & CFAR \\ 
       \hline
FedAvg    
    & 61.5 & \textbf{56.4} & 88.9 & 40.2 & 68.1  & \textbf{61.4}\\
    %\hline
FedProx
    & 58.4 & 50.1 & 85.4 & 35.6 & 66.8 & 60.5\\
    %\hline
MOON    
    & 57.7 &  48.0 & 86.9 & 37.6 & 66.2 & 58.6\\
FedAdam    
    & 49.4 & 16.9 & 81.0     & 28.9 & 54.0    & 40.5   \\
FedNova    
    & 58.4 & 47.1 & 83.6     & 37.0 & 64.3    & 60.3   \\
IST    
    & 64.3 & 50.8 & 90.7 & 39.8 & \textbf{69.5} & 49.4\\
    %\hline
ISTProx
    & \textbf{64.8} &  50.3 & \textbf{91.3} & \textbf{40.6} & 68.2 & 56.7\\
       \hline
\end{tabular}
\label{table:final_acc10}

    \vspace{10 pt}
    \centering
\begin{tabular}{cccccccc}
\multicolumn{7}{c}{(b) Communication (GB) to threshold acc.} \\
\hline
   & Birds & Cars & Flwrs & Aircrft & Textre & CFAR \\ 
       \hline
FedAvg    
    & 131 & \textbf{62} & 51 & 79        & 55 & 59 \\
    %\hline
FedProx
    & 250 & FAIL & 124      & FAIL        & 71 & \textbf{54}\\
     %  \hline
MOON    
    & FAIL & FAIL & 193      & FAIL       & 143 & 74  \\
FedAdam    
    & FAIL & FAIL & FAIL     & FAIL & FAIL    & FAIL   \\
FedNova    
    & 743 & FAIL & 430     & 608 & 309    & 160   \\
IST    
    & 79 & FAIL    & \textbf{37}  & \textbf{119}      & \textbf{69} & FAIL  \\
    %\hline
ISTProx
    & \textbf{72} & FAIL & 59     & 134        & 72 & 219 \\
       \hline
\end{tabular}

    \vspace{10 pt}  
    
\begin{tabular}{cccccccc}
\multicolumn{7}{c}{(c) GFLOPs to theshold acc} \\
\hline
   & Birds & Cars & Flwrs & Aircrft & Textre & CFAR \\ 
       \hline
FedAvg    
    & 13121 & \textbf{6245} & 5150 & 7892 & 5461 & 5884 \\
    %\hline
FedProx
    & 1000 & FAIL & 495  & FAIL & 283 & \textbf{218}\\
     %  \hline
MOON    
    & FAIL & FAIL & 581      & 1243 & 653 & 1207\\
FedAdam    
    & FAIL & FAIL & FAIL     & FAIL & FAIL    & FAIL   \\
FedNova    
    & 14872 & FAIL & 8614     & 12172 & 6181    & 3200   \\
IST    
    & 315 & FAIL & \textbf{148}      & \textbf{475} & \textbf{277}  & FAIL       \\
    %\hline
ISTProx
    & \textbf{287} & FAIL & 238      & 535 & 288  & 875      \\
       \hline
\end{tabular}

    \caption{Skewed data results, 100 sites, 10\% participation.}
    \label{fig-skewed}
\end{table}

\subsection{Tuning and Evaluation}

For a given method, using the Birds data set, we first perform a grid search over the various hyperparameters, and for each hyperparameter set, we run the method for a ``long time,'' where a ``long time'' allows a maximum of $10^{14}$ floating point operations (or 100 teraFLOPs, where floating point operations are summed across all devices participating in federated training) and five terabytes of data communicated (assuming single-precision floating point computations).  Note that 100 teraFLOPs corresponds to around 14 hours of CPU time on a single mobile phone, or perhaps several hundred hours of processor time on a limited IoT device.  Once a method exceeds 100 teraFLOPs or five terabytes of communication, we take its very best observed test accuracy, averaged over ten communication rounds.  The best average observed over all hyperparameter settings is taken as the method's ``final accuracy''.

\begin{figure*}
  \centering
  \begin{subfigure}{0.32\linewidth}
  \centering
    \includegraphics[width=1\textwidth]{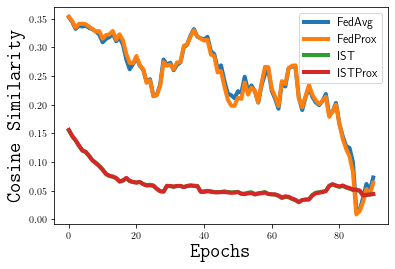}
    \caption{i.i.d., Cars, one round accuracy.}
  \end{subfigure}
  \hfill
    \begin{subfigure}{0.32\linewidth}
  \centering
    \includegraphics[width=1\textwidth]{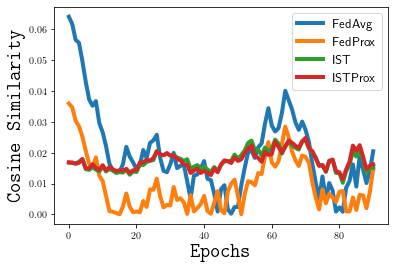}
    \caption{Skewed, Cars, one round accuracy.}
  \end{subfigure}
  \hfill
  \begin{subfigure}{0.32\linewidth}
  \centering
    \includegraphics[width=1\textwidth]{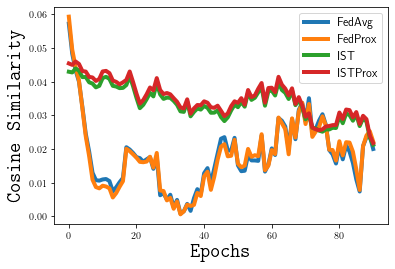}
    \caption{Skewed, Cars, 20 rounds accuracy.}
  \end{subfigure}

  \centering
  \begin{subfigure}{0.32\linewidth}
  \centering
    \includegraphics[width=1\textwidth]{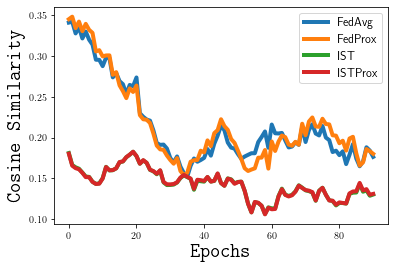}
    \caption{i.i.d., Flowers, one round accuracy.}
  \end{subfigure}
  \hfill
    \begin{subfigure}{0.32\linewidth}
  \centering
    \includegraphics[width=1\textwidth]{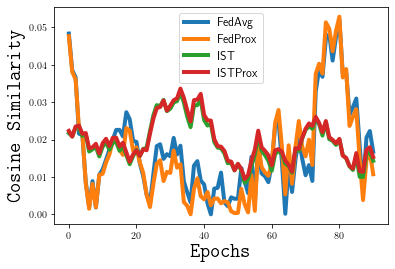}
    \caption{Skewed, Flowers, one round accuracy.}
  \end{subfigure}
  \hfill
  \begin{subfigure}{0.32\linewidth}
  \centering
    \includegraphics[width=1\textwidth]{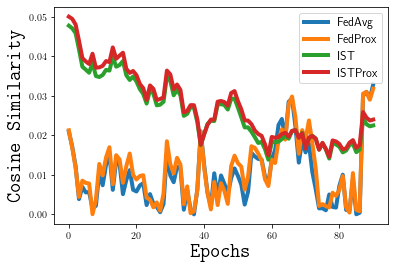}
    \caption{Skewed, Flowers, 20 rounds accuracy.}
  \end{subfigure}
  
 \caption{Comparison of similarity of the FL descent direction to a centrally-computed descent direction. One means the directions are identical; 0 represents de-correlation. At left is i.i.d. data; center is skewed data, similarity computed over one descent step; right is skewed data, similarity computed over 20 descent steps. }
  \label{fig:cosine}
\end{figure*}

Once we have the method's final accuracy, we then perform a grid search over all hyperparameters to find the two sets of parameter settings for the method that (a) minimize the number of FLOPs to reach 90\% of the final accuracy, and (b) minimize the number of GB of communication necessary to reach 90\% of the final accuracy.  Now we have three sets of hyperparameters: one optimized for accuracy, one for gigaFLOPs, and one for communication.  \textit{The reason that we use three sets of hyperparameters is that the goals of maximizing accuracy, minimizing FLOPs and minimizing communication are often in direct opposition to one another, and these resources can often be traded} (e.g., more local iterations and FLOPs for less communication).

Next, we consider each data set in sequence.  We first run each method to 100 teraFLOPs or five terabytes of communication on the data set, using the accuracy-optimized parameters.  We then compute the maximum accuracy that any method obtained on the data set, and set $90\%$ of this accuracy as a target.  Then, for each method and each data set, we compute two additional numbers: the number of gigaFLOPs required to reach the target accuracy for five consecutive rounds (using the FLOP-optimized settings), and the number of gigabytes of communication required to reach the target accuracy (using the communication-optimized settings).  If a method is unable to reach the target accuracy within the 100 teraFLOPs/five terabytes limit, it is said to have failed (indicated as FAIL in tables). 

We estimate the amount of FLOPs used during training using the FlopCountAnalysis tool in the Python \textit{fvcore} \cite{fvcore}. Similarly, for our communication costs we estimate the size of a model in bytes with the Python \textit{torchinfo} \cite{torchinfo} package.  

The reason for this evaluation strategy is simple.  In practice, the preferred FL method is likely to be the one that reaches a high level of accuracy with little cost.  Hence, we (somewhat arbitrarily, but precisely) define ``high level of accuracy'' as 90\% of the peak accuracy observed over all methods for a given data set.  Given this target, evaluation is then a matter of counting floating point operations and bytes to reach that target.  

Note that other metrics are possible: power usage and elapsed wall clock time are two of the most obvious.  However, both of these will be closely related to FLOPs and communication required, and hence we stick to these two.

\section{Identically Distributed Data} \label{sec:hypo_pretrain}
%%\subsection{Motivation}

%%\textcolor{red}{\textit (1) We want to compare the floating point and bytes transferred between full-training and a pre-trained backbone.} 

%%\textcolor{red}{\textit (2) We want to evaluate the performance of FL algorithms in the "easy" case of i.i.d. data.} 

\subsection{Setup}

In our first set of experiments, the images are randomly partitioned across the sites, so that each site has a randomly-selected (with replacement) subset of the overall set of images in the training data set.  After a round of communication finishes, ten sites are randomly selected as being ``on the network'' and those sites participate in the present round of training.  After local training finishes and all of the results have been communicated, another set of ten sites are randomly selected and the process continues.  

For this simple ``i.i.d.'' setup, our experiments will answer a few key questions.  First and foremost, should we be doing full backpropagation through a convolutional network, or should we be exclusively using a pre-trained backbone?  Second, when one considers computation and communication (rather than just final accuracy) do the methods compare differently than when the only metric is accuracy?  And finally, does the data set used matter?
Results are shown in Table \ref{fig-iid}. For each data set, the best result is indicated in bold.

\subsection{Discussion}

There are a few clear results.  First, except for FedAdam and FedFull, the methods generally have very similar final accuracy, with the two IST-based methods being slightly higher.  Thus, one key finding is that for data without skewness across sites, \emph{if one is only interested in final accuracy, it does not seem to matter much which method is used}, and this result holds across data sets.  The two exceptions to this are FedAdam and FedFull.  FedAdam does not reach the same final accuracy as the other methods, but it generally does quite a good job reaching high accuracy with relatively little communication.  This corresponds to the fact that it tends to converge in relatively few communication rounds.

FedFull is a very interesting case.  Note that it has poor final accuracy---only 3\% in the case of the  Cars data set.  As a result, it is labeled as ``FAIL'' in every case, because it can never reach 90\% of the peak accuracy observed for the other methods on a given data set.  One might ask, \textit{Why is FedFull such a poor choice?}  The answer is simple: \textit{as we put a 100 teraFLOP/five terabyte ``timeout'' on the computation, FedFull timed out in every case.} That is, it \emph{always} hit the 100 teraFLOP/five terabyte limit.  Were FedFull allowed infinite computation and infinite communication, it is actually the preferred method, as the pre-trained backbones limit accuracy.  For example, consider Figure \ref{fig:fedprox_full_vs_pre}, which plots accuracy as a function of FLOPs for both FedFull and FedProx on Cars.  FedFull eventually reaches 90\% accuracy, but it takes nearly \emph{four orders of magnitude} more FLOPs to top out compared to FedProx. This FLOP usage gap is apparent in \cref{fig:fedprox_full_vs_pre} where even the first communication round of FedFull requires more FLOPs than the amount required for the entire FedProx training. As a result, we would argue that full backpropagation is effectively unusable in resource-constrained FL.  
Note, however, that our FedFull was trained using an initialization consisting of random weights.  Perhaps a hybrid method makes sense, where one begins with FedProx (with a pre-trained backbone) and then once the MLP begins to reach peak accuracy, we allow backpropagation over the backbone to begin.  This issue deserves further investigation.  That said, given our maximum FLOP count and communication, FedFull is not competitive and in the interest of space, we will not consider it further.

The final observation we make is the general superiority of IST, over each of the six data sets.  It is perhaps not surprising that IST is efficient in terms of communication and in terms of computation.  Each site obtains only a fraction of the MLP that is being trained, saving communication \emph{and} computation at each site.  However, it is surprising that it is \emph{also} generally the most accurate method, in terms of final accuracy (or close to it) in every case.  We will examine this finding more fully, later in the paper.

\section{Effect of Data Skewness} \label{sec:hypo_noniid}

%%\subsection{Motivation}

%%\textcolor{red}{\textit (1) We are comparing the FL algorithms in a skewed data setting and determine how drastically different their performance is compared to the i.i.d. case.} 

%%\textcolor{red}{\textit (2) In the centralized learning correlation experiment, our goal is to evaluate how closely averaging versus decomposition algorithms align with centralized training.}  

\subsection{Setup}

How  might these results change if data are not uniformly distributed across sites? To answer this question, we modify our experiments as follows.  Rather than assigning each training image to each site with uniform probability, we sample the class probabilities for each site from a Dirichlet$(0.01)$ distribution, and rerun our experiments (with 100 sites and 10 sites active in each round).  This creates a relatively extreme---yet still realistic---amount of skew.  For the six data sets we tested, the three-tuple $($number of classes in data set, min number of classes at any site, max number of classes at any site$)$ is:  for Birds $($200, 7, 20$)$,  Aircraft $($100, 2, 15$)$, Stanford Cars $($196, 4, 20$)$,  Flowers $($102, 3, 15$)$, Textures $($47, 1, 9$)$, and CIFAR100 $($100, 1, 13$)$.  This matches the realistic case where each site has access to only a few of the classes, and from this limited local information, an accurate global model must be constructed.   Results are shown in Table \ref{fig-skewed}.

\begin{table}[t]
\centering
    
    \setlength\tabcolsep{3.7pt}
    
\begin{tabular}{cccccccc}
\multicolumn{7}{c}{(a) 1000 sites } \\
\multicolumn{7}{c}{Change in acc., 2\% $\rightarrow$ 6\% participation} \\
\hline
& Birds & Cars & Flwrs & Aircrft & Textre & CFAR \\ 
       \hline

FedAvg    
& 0.1	& \textcolor{red}{-0.8}	& 0	& 0	& \textcolor{red}{-1.1}	& 1.3\\
    %\hline
FedProx
& 0.1	& \textcolor{red}{-2}	& 0.4	& 0.8	& 1	& 1.1\\
    %\hline
MOON    
 & 0.3	& 2.9	& 0.5	& 0.5	& \textcolor{red}{-0.2}	& 1.4\\
FedAdam    
& 12.8	& 16.8	& 7	& 12.5	& 12.1	& 12.4\\
FedNova    
& 0.6	& \textcolor{red}{-0.5}	& \textcolor{red}{-0.5}	& 0	& 0.6	& 0.8\\
& 1.2	& 7.3	& 0.5	& 2.4	& 0.7	& 4.9\\
    %\hline
ISTProx
& 0.4	& 6.3	& 0.4	& 1.8	& 7.6	& 4.1\\
       \hline
\end{tabular}
\label{table:acc2to6}

    \vspace{10 pt}
        \setlength\tabcolsep{3.7pt}

    \begin{tabular}{cccccccc}
\multicolumn{7}{c}{(b) 100 sites } \\
\multicolumn{7}{c}{Change in acc., 10\% $\rightarrow$ 30\% participation} \\
\hline
 & Birds & Cars & Flwrs & Aircrft & Textre & CFAR \\ 
       \hline
FedAvg    
& \textcolor{red}{-0.5}	& \textcolor{red}{-0.8}	& 0.6	& \textcolor{red}{-0.1}	& 1.2	& 4\\
    %\hline
FedProx
& \textcolor{red}{-0.4}	& \textcolor{red}{-2}	& 3.7	& 2.1	& \textcolor{red}{-0.6}	& 0\\
    %\hline
MOON    
& 1.5	& 2.9	& \textcolor{red}{-0.5}	& \textcolor{red}{-0.3}	& \textcolor{red}{-0.1}	& 2.8\\
FedAdam    
& 2	& 14.9	& \textcolor{red}{-0.9}	& 0.6	& 0.4	& 5\\
FedNova    
& 0.3	& \textcolor{red}{-0.5}	& 4.3	& \textcolor{red}{-0.7}	& \textcolor{red}{-0.6}	& \textcolor{red}{-0.3}\\
IST    
& 2.3	& 7.3	& \textcolor{red}{-1.7}	& 0.1	& 0.1	& 6.2\\
    %\hline
ISTProx
& 3	& 6.3	& \textcolor{red}{-0.9}	& 3	& 1.3	& 2.2\\
       \hline
\end{tabular}
\label{table:acc2to6}

    \caption{Observed change in accuracy, computed as ``(\% correct with fewer sites) - (\% correct with more sites)'', when active sites increases. \textcolor{red}{Red} indicates negative change.}
    \label{fig-acc-diff-sites}
\end{table}

\begin{table}[t]
\centering
    
    \setlength\tabcolsep{3.2pt}
    
\begin{tabular}{cccccccc}
\multicolumn{7}{c}{(a) 1000 sites } \\
\multicolumn{7}{c}{Change in comm., 2\% $\rightarrow$ 6\% participation} \\
\hline
 & Birds & Cars & Flwrs & Aircrft & Textre & CFAR \\ 
       \hline
FedAvg    
& \textcolor{red}{3.02}	& 2.41	& 2.53	& \textcolor{red}{4.28}	& 2.25	& 2.07\\
    %\hline
FedProx
& 2.76	& NA	& 2.34	& 2.98	& 2	& 2.12\\
    %\hline
MOON    
& 2.78	& 0.88	& 2.52	& 2.91	& 2.28	& 2.11\\
FedAdam    
& NA	& NA	& 0	& NA	& NA	& NA\\
FedNova    
& 2.6	& \textcolor{red}{3.81}	& 2.43	& \textcolor{red}{3.73}	& 2.77	& \textcolor{red}{3.12}\\
IST    
& 1.41	& 0	& 1.07	& 1.95	& 1.03	& NA\\
    %\hline
ISTProx
& 1.29	& NA	& 1.15	& 2.2	& 0.48	& NA\\
       \hline
\end{tabular}
\label{table:acc2to6}  
    
    \vspace{10 pt}
    
\begin{tabular}{cccccccc}
\multicolumn{7}{c}{(b) 100 sites } \\
\multicolumn{7}{c}{Change in comm., 10\% $\rightarrow$ 30\% participation} \\
\hline
 & Birds & Cars & Flwrs & Aircrft & Textre & CFAR \\ 
       \hline
FedAvg    
& 0.65	& NA	& 1.97	& 0.24	& 1.24	& 0\\
    %\hline
FedProx
& \textcolor{red}{$\infty$}	& \textcolor{red}{3.13}	& 2.82	& \textcolor{red}{6.43}	& 1.75	& 2.41\\
    %\hline
MOON    
& \textcolor{red}{$\infty$}	& NA	& 1.27	& NA	& \textcolor{red}{5.31}	& \textcolor{red}{7.39}\\
FedAdam    
& NA	& NA	& 1.29	& NA	& 2.23	& 2.89\\
FedNova    
& 0.3	& \textcolor{red}{-0.5}	& 4.3	& \textcolor{red}{-0.7}	& \textcolor{red}{-0.6}	& \textcolor{red}{-0.3}\\
IST    
& \textcolor{red}{$\infty$}	& NA	& 1.09	& \textcolor{red}{$\infty$}	& 2.73	& \textcolor{red}{6.44}\\
    %\hline
ISTProx
& NA	& NA	& NA	& NA	& NA	& NA\\
       \hline
\end{tabular}
\label{table:acc2to6}

    \caption{Multiplicative change in communication when tripling the number of active sites.  \textcolor{red}{Red} text indicates more than a $3\times$ increase; black indicates less than a $3\times$ increase. ``NA'' indicates the case when there was a FAIL in both bases. \textcolor{red}{$\infty$} indicates a change from non-fail to a FAIL, and 0 indicates a change from FAIL to non-fail. }
    \label{fig-comm-diff-sites}
\end{table}

\subsection{Discussion}

Probably the single most interesting finding is that for a highly-skewed situation, there seems to be divergence in performance, depending upon the data set. 
 The two IST methods seem to perform best overall, with either IST or ISTProx being the superior option in terms of final accuracy, FLOPs, or communication for four of the six data sets.  However,  IST does relatively poorly on the venerable CFAR data set.  CFAR  is a somewhat strange benchmark, as the images are tiny (32 $\times$ 32) and widely varied within the data set, and both IST variants have problems with this.  Nevertheless, it is a good stress test. Note that in the two cases where IST does not do the best, it is either simple FedAvg or FedProx that is the best performer.  One of the most surprising findings is that FedAvg dominates all other methods in all metrics (accuracy, FLOPs, communication) on  Cars.  
 
 Overall, across the two experiments (i.i.d. and skewed), we find that adding a proximal term to FedAvg (to obtain FedProx) seems to have little effect on accuracy, but \emph{does} speedup convergence, both in terms of reducing communication and FLOPs.  However, IST seems to be the best performer overall, allowing for significant reductions in communication and FLOPs.  The addition of a proximal term \emph{does not} seem to help IST much.  The other methods tested all seem to under-perform.

\subsection{FL, Approximating Centralized Learning}

How does the training process for a method (such as FedAvg and FedProx) that trains the entire model at each site and then reconciles the various versions of the model, compare with IST, which avoids this through decomposition? 

To examine, we centrally train a model from initialization through convergence using min-batch gradient descent, obtaining model parameters $M_i$ for the $i$th training epoch.  Then, using FL, we start training in a federated setting from $M_i$ for some number of communication rounds, to obtain new  model parameters $\hat{M}_i$.  We refer to the set of model parameters we \emph{would} have obtained after processing exactly the same set of data centrally, via a series of mini-batches, as $M_i^*$.  Then the cosine similarity of the directions both methods move, defined as $(\hat{M}_i - M_i) \cdot (M_i^* - M_i) / (||\hat{M}_i - M_i|| \times ||M_i^* - M_i||)$, shows the utility of FL as an approximation for centralized learning. Cosine similarity close to one means FL and centralized training move in the same direction.  Zero implies de-correlation.

Some results obtained in this way are plotted in Figure \ref{fig:cosine}, where we consider FedAvg, FedProx, IST, and ISTProx.  These plots depict this cosine similarity as a function of the training  epochs.  We plot running averages to smooth the data. The top three plots are for Cars, the bottom three Flwrs.  The results for identically distributed data are at left, whereas skewed data are in the center, and at the right. 

FedAvg and FedProx are shown to approximate the centralized gradient descent direction well early on, but the similarity drops after initial training.  This  makes sense, as initially, with a random initialization, learning is ``easy'' as the model is far from optimal.  In the i.i.d. case FedAvg and FedProx perform well.  But in the skewed case, after the initial training period, similarity to the centralized training direction drops. IST seems more well-behaved, with less variance. Especially considering
how well the methods approximate the direction of centralized training through twenty rounds on skewed data (the right column),
the IST-based methods seem to do better. This is especially pronounced for the Flwrs data set through the first 20 epochs, where IST averages a similarity of 0.035, compared to 0.01 for the averaging-based methods. Perhaps this is due to the high variance of the directions computed by FedAvg and FedProx---as estimators for the centrally-predicted direction---compared to IST.  Examining why this is deserves further investigation in future work.

%Most remarkable is that when we test how well the methods approximate the direction of centralized training through twenty rounds (the right column), the IST-based methods dominate. The way to interpret this is that in the skewed case, decomposition has significantly higher similarity with centralized training long-term, compared to reconciliation.  This is especially pronounced for the Flwrs data set (where IST seems to do much better than the averaging approaches; see Figure \ref{fig-skewed}). Perhaps this is due to the high variance of the directions computed by FedAvg and FedProx---as estimators for the centrally-predicted direction---compared to IST.  Examining why this is deserves further investigation in future work.

\section{Varying Sites and Connectivity} 

%%\subsection{Motivation}

%%\textcolor{red}{\textit (1) We want to determine the scalability of the different FL algorithms for different rates and numbers of client participation. The aim is to determine how both averaging and decomposition fare when the FL training scheme involves more clients.} 

\subsection{Setup}

All of our experiments thus far have focused on the simple case of 100 sites, where at each communication round, 10\% of the sites are connected.  We now ask: how do things change when the number of sites changes and/or the number of sites connected at each communication round changes?  Keeping all of the other settings the same, we try 1000 sites, with either 2\% or 6\% of the sites connected at each communication round, and 100 sites, with either 10\% of 30\% of the sites connected at each round.  Due to space constraints, we consider accuracy and communication here, and provide further results of these experiments in the appendix.  In Table \ref{fig-acc-diff-sites}, we show the absolute change in percentage accuracy moving from 2\% to 6\% connectivity, and from 10\% to 30\% connectivity.  In Table \ref{fig-comm-diff-sites}, we show the multiplicative change in communication required.  Note that, as we increase the number of active sites by $3\times$, we might expect a $3\times$ increase in communication if the same number of communication rounds are required.  If fewer rounds are required, we would see an increase that is less than $3\times$.

\subsection{Discussion}

  FedAdam, IST, and ISTProx are generally helped a bit by increasing participation.  But for the other methods, the results obtained by increasing participation rates are mixed, with a surprising number of red values in both Table \ref{fig-acc-diff-sites} and Table \ref{fig-comm-diff-sites}, indicating a decrease in accuracy or communication efficiency via the addition of extra active sites.  In addition, it seems that the increase from 2\% participation to 6\% participation is much more helpful than 10\% to 30\%.

The lack of clear benefit to increasing the rate of participation is not surprising in retrospect.  Aside from FedFull, each of the methods was able to run to convergence in the allotted FLOP/communication budget.  The central question becomes: can we run to convergence in fewer rounds by doing more work in each round?  This seems very unlikely, as this is akin to increasing the batch size in centralized learning, and expecting a significant decrease in the number of batches required for convergence.  There is typically \emph{some} decrease in the number of batches with increasing batch size, but it often is not significant.  In reconciliation-based methods, this may be particularly inefficient as local training rounds cause the active sites to diverge and simply makes it more difficult to reconcile them. IST may have another advantage in that regard, in that more active sites does not increase communication for IST, it simply means that each site gets less of the full model.  As long as each site can tolerate local learning using a smaller model (due to partitioning more ways), the learning may benefit from seeing a greater variety of data in each round.  This certainly seems to be the case in when moving from 2\% to 6\%, which benefits IST.

\section{Conclusion}

We have designed a set of experiments to evaluate FL methods.  Among the issues considered were what were the appropriate evaluation metrics (accuracy alone makes little sense), whether to use a pre-trained backbone, and the effect of connectivity on the efficiency of the learning.  We find that overall, the method of decomposition of a neural network into independent subnetworks seems to be the best option.  This has the benefit of decreasing communication and computation compared to ``reconciliation-based'' methods that train a copy of the model at each site, as the model is sharded, rather than broadcasted. 

\vspace{5 pt}
\noindent
\textbf{Acknowledgements.} We would like thank the anonymous reviewers for their comments on the submitted version of the paper. Work presented in this paper has
been supported by an NIH CTSA, award No. UL1TR003167 and by
the NSF under grant Nos. 1918651, 1910803, 2008240, and 2131294.

{\small
\bibliographystyle{ieee_fullname}
\bibliography{egbib}
}

\newpage
\clearpage
\newpage

\appendix

\vspace{5 pt}
\noindent
\textbf{\Large Appendix}

\section{Data Processing Details}

To reiterate from the main paper, the datasets we used in our experiments are CUB-200-2011
(Birds) \cite{birds}, Stanford Cars (Cars) \cite{cars}, (3) VGGFlowers
(Flwrs) \cite{flowers}, (4) Aircraft (Aircrft) \cite{aircraft}, (5) Describable
Textures (Textre) \cite{dtextures}, and (6) CIFAR 100 (CFAR) \cite{cifar}. All datasets are partitioned along their original train-test split. For datasets with an additional validation set, in the case of Textre, Aircrft, and Flwrs, we merge this with the test dataset. The Aircrft dataset has different levels of aircraft type - Manufacturer, family, variant. Variant is the specific model of aircraft and we chose this as the label for highest granularity. 

On the Birds, Flwrs, Aircrft, and Textre datasets we perform a randomized cropping followed by a $224\times 224$ resize on the train datasets, then a randomized horizontal flip, and a normalization of the RGB colors. On the test datasets we resize the images to $256\times 256$ and then center crop them to $224\times 224$, and again normalize the colors. For Cars and CFAR datasets we use the same processing we used for the test datasets for both the train and test datasets. These images are then converted to tensors to be used as inputs. 

For inputs to all algorithms except FedFull, we input these image tensors into pretrained ResNet101 and a DenseNet121 convolutional neural networks. We use the existing versions downloaded from PyTorch, which are trained on the ImageNet dataset. We remove the final linear layer from these CNNs to retrieve the feature tensors as outputs which we concatenate and use as the inputs of most of the evaluated algorithms. ResNet101 produces 1024-dimensional vectors while DenseNet121 produces ones of dimension 2048, so the concatenated feature vector is 3072-dimensional. 

To sample the data for $N$ client devices and a dataset with $k$ classes, we generate $N$ vectors of length $k$, one for each client, drawn from a Dirichlet distribution. More precisely, given concentration parameter $\alpha$ and for a task of $k$ classes, we sample client class distribution vectors $(x_1,\ldots,x_k)\in [0, 1]^k$, such that $\sum_{i=1}^k x_i=1$, according to the following probability density function:
\begin{align}   
    p_{\alpha}(x_1,\ldots,x_k) = \frac{1}{B(\alpha)}\prod_{i=1}^{k}x_{i}^{\alpha - 1}\label{diric_def}
\end{align}

where $B(\alpha)$ is a normalization. The concentration parameter $\alpha$ dictates the skewness of data with values closer to $0$ being more skewed and larger values generating more i.i.d. samples. More generally the Dirichlet concentration parameter can be assigned per class by replacing $\alpha$ with $\alpha_i$ in \cref{diric_def}, but we use the same value on all classes in the generated client distributions. 

For each client we assign 500 images by sampling from each class, with replacement, the proportion given by these generated distribution vectors. The sampling is done with replacement because in many datasets some classes may not have sufficiently many examples, especially in the non-i.i.d. case.

\section{Simulation Implementation Details}

All of our experiments are done as simulations of synchronous, centralized federated learning. We assume a central server that broadcasts a global model that client devices download and upload for their local training. Our simulator is implemented with PyTorch \cite{pytorch} 

Prior to training we initialize the client data distributions via Dirichlet sampling and these samples are fixed for an entire simulation. At the start of a global communication and training round, we randomly select the subset of clients that participate in the training. For IST and ISTProx this is also when the model partitioning is done. 

In the centralized learning approximation experiments, we also train the central MLP models on the central server without partitioning to clients. In this case, for consistency of training data, we concatenated the sampled datasets of the participating clients and used this as the training set. 

\subsection{Models}

Except FedFull, all algorithms use a single-hidden layer multilayer perceptron (MLP) as their base model. This consists of linear layer, followed by a BatchNorm (Batch Normalization) layer then a ReLU activation function, and then another linear layer followed by a softmax. As our pretrained extractor outputs are 3072-dimensional vectors, the input layer of the MLPs are 3072 wide, while the output layer has the number of classes of a particular dataset, e.g. on Birds we would have 200 width output layer.  

For all algorithms except for IST, ISTProx, and FedFull, the MLP hidden layer has 1000 neurons. FedFull uses the ResNet18, initialized with the default weights pretrained on the ImageNet dataset and then fine-tuned on the various datasets. 

For IST and ISTProx, we find scaling the hidden neurons based on the number of concurrent training sites made sense: 3000 neurons for ten sites, 6000 for 20, 9000 for 30, 18000 for 60, and so on. This ensures that the client models are not too small and in this setup the IST client models share the MLP architecture but with 300 hidden neurons in all cases. 

For MOON, we also extract the output after the ReLU activation function as the learned representation used in its contrastive term. 

\subsection{Training}

We use batch size of 32 for all datasets and methods. A single local training round computes one batch. A learning rate of 0.01 is chosen on all algorithms except FedAdam which uses 0.03, with a 10x decay after 50\% and 75\% of total communication rounds have passed. In the centralized training case, we instead train an entire epoch by passing through all the concatenated data samples once. 

Prior to training, we generate estimations of the FLOP usage of a forward or backward pass per batch via the Python \textit{fvcore} \cite{fvcore} package's FLOPCountAnalysis as well as the model size via the \textit{torchinfo} \cite{torchinfo} package. After a communication round, we estimate the total bytes transferred by multiplying the model size by the number of participating clients and then double this to account for both downloading and uploading the model to the central server. To compute the FLOP usage after a communication round, the FLOPCountAnalysis gives an estimate of the forward propagation that we multiply by two to account for the backward pass and then by the number of local rounds and the number of participating clients. For MOON this FLOP estimate is also doubled to account for its extra two forward pass computations. 

Except FedProx, ISTProx, and FedNova, all methods use a standard stochastic gradient descent optimizer for the local training with a momentum weight of 0.9. After a global communication round, we reinitialize these optimizers to avoid reusing stale momentum from the previous training. 

FedProx and ISTProx use a proximal optimizer that also takes as input the initial state of the model before local training is done. 

FedNova uses its own optimizer that is based on an implementation from \cite{fednova}. 

\subsection{Testing and Evaluation}

As our goal is to evaluate the generalizability of the aggregated model, the testing dataset is taken as a whole rather than sampled by a Dirichlet distribution as in the case of the training set. Our final accuracy is computed by the proportion of the top-1 predicted classes that match the true label. 

These methods may have inconsistent accuracy performance within a few communication rounds. So in computing the final accuracies, we average over a sliding window of 10 communication rounds to smooth out the accuracies. 

We reiterate the process detailed in the main paper that we choose the highest accuracy among all algorithms and then take 90\% of this accuracy as our threshold. We then identify the smallest GB transferred and GFLOP count needed to reach this 90\% threshold.

\begin{table}[t]
\centering
    
    \setlength\tabcolsep{3.2pt}
    
\begin{tabular}{cccccccc}
&\multicolumn{6}{c}{(a) Final accuracies} \\
\hline
   & Birds & Cars & Flwrs & Aircrft & Textre & CFAR \\ 
       \hline
FedAvg    
    & 61.0 & 55.6 & 89.5 & 40.1 & 69.3 & \textbf{65.4} \\
    %\hline
FedProx
    & 58.0 & 48.1 & 89.1 & 37.7 & 66.2 & 60.5\\
   %\hline
MOON    
    & 59.2 & 50.9 & 86.4 & 37.3 & 66.1 & 61.4 \\
FedAdam    
    & 51.4 & 31.8 & 80.1     & 29.5 & 54.4    & 45.5   \\
FedNova    
    & 58.7 & 46.6 & 87.9     & 36.3 & 63.7    & 60.0   \\
FedFull     
    & 1.5 & 0.9 & 3.0     & 1.6 & 4.1    & 1.2   \\
    %\hline
IST    
    & 66.6 & \textbf{58.1} & 89.0 & 39.9 & \textbf{69.6} & 55.6\\
    %\hline
ISTProx
    & \textbf{67.8} & 56.6 & \textbf{90.4} & \textbf{43.6} & 69.5 & 58.9\\
       \hline
\end{tabular}

    \vspace{10 pt}
    
\begin{tabular}{cccccccc}
&\multicolumn{6}{c}{(b) Communication (GB) to threshold acc.} \\
\hline
   & Birds & Cars & Flwrs & Aircrft & Textre & CFAR \\ 
       \hline
FedAvg    
     & FAIL & \textbf{194} & 144 & 508        & \textbf{96} & \textbf{142} \\
    %\hline
FedProx
    & FAIL & FAIL & 157      & FAIL       & 377 & 399\\
     %  \hline
MOON    
    & FAIL & FAIL & 249    & FAIL       & 319 & 214  \\
FedAdam    
    & FAIL & FAIL & FAIL     & FAIL & FAIL    & FAIL   \\
FedNova    
    & FAIL & FAIL & 469     & FAIL & 843    & 1031   \\
FedFull     
    & FAIL & FAIL & FAIL     & FAIL & FAIL    & FAIL   \\
    %\hline
IST    
    & 183 & 279     & \textbf{129}      & 644      & 115 & FAIL  \\
    % \hline
ISProx
    & \textbf{150} & FAIL & 148      & \textbf{318}    & 143 & FAIL\\
       \hline
\end{tabular}

    \vspace{10 pt}  
    \centering
\begin{tabular}{cccccccc}
&\multicolumn{6}{c}{(c) GFLOPs to threshold acc.} \\
\hline
   & Birds & Cars & Flwrs & Aircrft & Textre & CFAR \\ 
       \hline
FedAvg    
    &  FAIL & FAIL & 7057 & FAIL & 6222 & FAIL \\
    %\hline
FedProx
    & FAIL & FAIL & 630      & FAIL & 1507 & \textbf{1597}\\
     %  \hline
MOON    
    & FAIL & FAIL & 1321     & FAIL & 1359    & 1815   \\
FedAdam    
    & FAIL & FAIL & FAIL     & FAIL & FAIL    & FAIL   \\
FedNova    
    & FAIL & FAIL & 9392     & FAIL & 16877    & 20625   \\
FedFull     
    & FAIL & FAIL & FAIL     & FAIL & FAIL    & FAIL   \\
IST    
    &  1400 & FAIL  & \textbf{367}     & 2610 & 754  & FAIL     \\
    %\hline
ISTProx
    & \textbf{602} & FAIL & 594      & \textbf{1274} & \textbf{572}    & FAIL    \\
       \hline
\end{tabular}
    \caption{Skewed data results, 100 sites, 30\% participation.}
    \label{fig-skew-30}
\end{table}

\begin{table}[t]
\centering
    
    \setlength\tabcolsep{3.2pt}
    
\centering
\begin{tabular}{cccccccc}
&\multicolumn{6}{c}{(a) Final accuracies} \\
\hline
   & Birds & Cars & Flwrs & Aircrft & Textre & CFAR \\ 
       \hline
FedAvg    
    & 71.5 & \textbf{56.4} & 95.6 & 46.2 & 73.0  & 72.7\\
    %\hline
FedProx
    & 70.2 & 50.1 & 95.4 & 45.7 & 71.9 & 71.7\\
    %\hline
MOON    
    & 70.3 &  48.0 & 95.1 & 45.7 & 71.7 & 71.5\\
FedAdam    
    & 47.2 & 28.2 & 82.5     & 22.7 & 48.2    & 42.6   \\
FedNova    
    & 70.8 & 47.1 & 95.7     & 46.2 & 71.9    & \textbf{74.1}   \\
FedFull     
    & 1.0 & 0.7 & 1.7     & 1.3 & 3.3    & 1.3   \\
    %\hline
IST    
    & 73.3 & 50.8 & 95.8 & 46.8 & \textbf{75.0} & 59.9\\
    %\hline
ISTProx
    & \textbf{73.7} &  50.3 & \textbf{96.1} & \textbf{48.1} & 68.2 & 61.7\\
       \hline
\end{tabular}

    \vspace{10 pt}
    
\centering
\begin{tabular}{cccccccc}
&\multicolumn{6}{c}{(b) Communication (GB) to threshold acc.} \\
\hline
   & Birds & Cars & Flwrs & Aircrft & Textre & CFAR \\ 
       \hline
FedAvg    
     & 122 & \textbf{74} & 36 & \textbf{162} & \textbf{65} & \textbf{92} \\
    %\hline
FedProx
    & 141 & FAIL & 44 & 191 & 79 & 128\\
     %  \hline
MOON    
    & 137 & 208 & 44    & 220  & 69 & 129  \\
FedAdam    
    & FAIL & FAIL & FAIL     & FAIL & FAIL    & FAIL   \\
FedNova    
    & 377 & 478 & 113     & 465 & 146    & 234   \\
FedFull     
    & FAIL & FAIL & FAIL     & FAIL & FAIL    & FAIL   \\
    %\hline
IST    
    & 92 & FAIL & 29 & 343  & 133 & FAIL  \\
    % \hline
ISTProx
    & \textbf{85} & FAIL & \textbf{26}      & 254    & 122 & FAIL\\
       \hline
\end{tabular}

    \vspace{10 pt}  
    \centering
\begin{tabular}{cccccccc}
&\multicolumn{6}{c}{(c) GFLOPs to threshold acc.} \\
\hline
   & Birds & Cars & Flwrs & Aircrft & Textre & CFAR \\ 
       \hline
FedAvg    
    & 12243 & \textbf{7410} & 3586 & 16190 & 6523 & 9175 \\
    %\hline
FedProx
    & 563 & FAIL & 176  & \textbf{766} & \textbf{317} & \textbf{513}\\
     %  \hline
MOON    
    & 1239 & FAIL & 380  & 1275 & 546 & 917\\
FedAdam    
    & FAIL & FAIL & FAIL     & FAIL & FAIL    & FAIL   \\
FedNova    
    & 7534 & 9568 & 2263     & 9307 & 2924    & 4682   \\
FedFull     
    & FAIL & FAIL & FAIL     & FAIL & FAIL    & FAIL   \\
IST    
    & 366 & FAIL & 115  & 1372 & 532  & FAIL       \\
    %\hline
ISTProx
    & \textbf{342} & FAIL & \textbf{104}      & 1018 & 487  & FAIL      \\
       \hline
\end{tabular}

    \caption{Skewed data results, 1000 sites, 2\% participation.}
    \label{fig-skew-2}
\end{table}

\begin{table}[t]
\centering
    
    \setlength\tabcolsep{3.2pt}
    
\begin{tabular}{cccccccc}
&\multicolumn{6}{c}{(a) Final accuracies} \\
\hline
   & Birds & Cars & Flwrs & Aircrft & Textre & CFAR \\ 
       \hline
FedAvg    
    & 71.6 & 55.6 & 95.6 & 46.2 & 71.9  & 74.0\\
    %\hline
FedProx
    & 70.3 & 48.1 & 95.8 & 46.5 & 72.9 & 72.8\\
    %\hline
MOON    
    & 70.6 &  50.9 & 95.6 & 46.2 & 71.5 & 72.9\\
FedAdam    
    & 60.0 & 45.0 & 89.5     & 35.2 & 60.3    & 55.0   \\
FedNova    
    & 71.4 & 46.6 & 95.2     & 46.2 & 72.5 & \textbf{74.9}   \\
FedFull     
    & 0.9 & 0.7 & 1.0     & 1.0 & 6.5    & 1.5   \\
    %\hline
IST    
    & \textbf{74.5} & \textbf{58.1} & 96.3 & 49.2 & 75.7 & 64.8\\
    %\hline
ISTProx
    & 74.1 &  56.6 & \textbf{96.5} & \textbf{49.9} & \textbf{75.8} & 65.8\\
       \hline
\end{tabular}

    \vspace{10 pt}
    
\centering
\begin{tabular}{cccccccc}
&\multicolumn{6}{c}{(b) Communication (GB) to threshold acc.} \\
\hline
   & Birds & Cars & Flwrs & Aircrft & Textre & CFAR \\ 
       \hline
FedAvg    
     & 369 & \textbf{178} & 91 & 693 & 146 & 190 \\
    %\hline
FedProx
    & 389 & FAIL & 103 & 569 & 158 & \textbf{271}\\
     %  \hline
MOON    
    & 381 & 183 & 111    & 641 & 157 & 272  \\
FedAdam    
    & FAIL & FAIL & 213     & FAIL & FAIL    & FAIL   \\
FedNova    
    & 982 & 1821 & 275     & 1735 & 405    & 729   \\
FedFull     
    & FAIL & FAIL & FAIL     & FAIL & FAIL    & FAIL   \\
    %\hline
IST    
    & 130 & 411 & 31  & 670  & 137 & FAIL  \\
    % \hline
ISTProx
    & \textbf{110} & FAIL & \textbf{30}      & \textbf{558}    & \textbf{58} & FAIL\\
       \hline
\end{tabular}

    \vspace{10 pt}  
    \centering
\begin{tabular}{cccccccc}
&\multicolumn{6}{c}{(c) GFLOPs to threshold acc.} \\
\hline
   & Birds & Cars & Flwrs & Aircrft & Textre & CFAR \\ 
       \hline
FedAvg    
    & 11958 & FAIL & 4141 & FAIL & 4921 & 8811 \\
    %\hline
FedProx
    & 1557 & FAIL & 412  & 2278 & 633 & \textbf{1083}\\
     %  \hline
MOON    
    & 3152 & FAIL & 751 & 5752 & 1422 & 2336\\
FedAdam    
    & FAIL & FAIL & 5875     & FAIL & FAIL    & FAIL   \\
FedNova    
    & 19646 & \textbf{36432} & 5509     & 34723 & 8112    & 14594   \\
FedFull     
    & FAIL & FAIL & FAIL     & FAIL & FAIL    & FAIL   \\
IST    
    & 569 & FAIL & 136 & 2495 & \textbf{231}  & FAIL       \\
    %\hline
ISTProx
    & \textbf{441} & FAIL & \textbf{120}      & \textbf{2232} & 233  & FAIL      \\
\hline
\end{tabular}
    \caption{Skewed data results, 1000 sites, 6\% participation.}
    \label{fig-skew-6}
\end{table}

\section{Additional Tables}
We include the following tables for additional results of the experiments in the main paper:

\begin{itemize}
    \item Skewed data, 100 sites, 30\% client participation \cref{fig-skew-30} 
    \item Skewed data, 1000 sites, 2\% client participation \cref{fig-skew-2} 
    \item Skewed data, 1000 sites, 6\% client participation \cref{fig-skew-6} 
\end{itemize}

These tables generally follow the same trends that we observed in the skewed 10\% participation experiments in the main paper. Again, we see that IST methods perform best across all three metrics on the Birds, Flwrs, Aircrft, and Textres dataset, but is inconsistent on the Cars and CFAR datasets. Averaging algorithms, in particular FedAvg, fare better on these two datasets. As we also addressed in the main paper, increasing client participation does not consistently nor significantly increase accuracies and generally increases the GFLOP and GB costs. 

However, we note that moving from 100 clients to the 1000 client population cases, the final accuracies greatly increase across all algorithms and datasets, and are more closely aligned with the final accuracies of the i.i.d. cases. This effect merits further investigation, although we believe this phenomenon is due to the increased population mitigating the effects of non-i.i.d. data.

\section{Hyperparameter Tuning}

All hyperparameter tuning was done on the Birds dataset and in the interest of \textit{off-shelf performance} we maintained the same sets of parameters for all other datasets. We have two separate sets of parameters for optimal communication and for FLOP usage due to potential trade-offs when optimizing for one over the other, although in practice we find many cases where \textit{one set of parameters work best for both}. We consider the number of local rounds of stochastic gradient descent as a hyperparameter with either 1, 5, or 25 iterations. Other hyperparameters are algorithmic specific. Unless otherwise stated, we selected the parameters by table-scanning all combinations of possible hyperparameter configurations and optimizing on each of FLOP and GB transferred. We detail each method and the choices of hyperparameters below:

\subsection{FedAvg}

FedAvg does not have any other hyperparameter besides local training rounds. We find that FedAvg performs better on both communication and FLOP usage with \textbf{25} local iterations. 

\subsection{FedProx}

FedProx has proximal hyperparameter $\mu \in [0.05, 0.1, 0.15, 0.2]$. For optimizing both FLOP and communication we find that $\mathbf{\mu=0.2}$ and \textbf{1} local round of training does best on both metrics. 

\subsection{FedFull}

We reuse the choice of $\mathbf{\mu=0.2}$ from FedProx, but we find that 1 local iteration no longer works due to training the full model that a single batch is insufficient. We chose \textbf{5} local iterations for all cases as we found 25 too slow, but in either case the method generally times out due to a single communication round requiring too many FLOPs. 

\subsection{MOON}

MOON has a weight hyperparameter for its contrastive term, similar to that of the proximal term used in FedProx, ISTProx, and FedNova. However this hyperparmater has range $\mu\in [1, 5, 10]$. The contrastive term also has a temperature parameter $\tau\in [0.1, 0.5, 1]$. 

For FLOP optimized run, we choose with \textbf{1} local round, $\mathbf{\mu=1, \tau=0.5}$

For communication optimized run, we choose with \textbf{5} local rounds, $\mathbf{\mu=10, \tau=0.1}$

\subsection{FedNova}

FedNova also uses proximal hyperparameter $\mu \in [0.05, 0.1, 0.15, 0.2]$ and we reuse the choice $\mathbf{\mu=0.2}$ along with a choice of \textbf{5} local training rounds. Additionally, FedNova has a parameter $\tau_{eff}$, the effective number of local training rounds. We use the recommended choice for $\tau_{eff}$ in \cite{fednova}, in the case of a nonzero proximal term, of dividing our local rounds by the number of participating clients. 

\subsection{FedAdam}

In the Adam algorithm of FedAdam, we use the standard recommended settings \cite{fedopt, adam}
of $\mathbf{\beta_1=0.9,\beta_2=0.99}$ for the momentum and RMSProp weight decays. We chose the adaptability parameter $\mathbf{\tau=0.01}$ from a range of $\tau\in[10^{-2}, 10^{-3}, 10^{-4}, 10^{-5}, 10^{-6}]$. We chose a local learning rate of \textbf{0.03} and global learning rate of \textbf{0.01} both from a range $[0.1, 0.03, 0.01, 0.003]$. For FLOP optimized FedAdam, we chose local rounds of \textbf{5}. For communication optimized, we chose local rounds of \textbf{25}

\subsection{IST}

IST has the local model size as a parameter. Here we tuned the central model size in the case of 10 client devices from [1000, 1500, 2000, 3000] and found that 3000 had the best final accuracy. Therefore we chose \textbf{300} as the local model size. Unlike other hyperparameters, we did not tune this to optimize for FLOPs or bytes. We found that \textbf{1} local round has best performance on both FLOPs and bytes transferred. 

\subsection{ISTProx}

We reuse the \textbf{300} local model size in IST. Like FedProx, ISTProx has a proximal term $\mu \in [0.05, 0.1, 0.15, 0.2]$. We did not see much of an impact with the proximal term on IST as it does with FedProx, but $\mathbf{\mu=0.2}$ is still consistently the best choice on both metrics in our tuning along with training on \textbf{1} local round. 

\end{document}